\def\paperID{15}
\def\confName{CVPR}
\def\confYear{2024}

\def\authorBlock{
    Jinshu Chen \qquad
    Bingchuan Li\footnotemark[1]\thanks{Corresponding author} \qquad
    Miao Hua \qquad
    Panpan Xu \qquad
    Qian He\\
    ByteDance Inc. \\
    {\tt\small \{chenjinshu, libingchuan, huamiao, xupanpan, heqian\}@bytedance.com}
}

\newif\ifreview 
\newif\ifarxiv 
\newif\ifcamera \newcommand{\cameraready}{\cameratrue}
\newif\ifrebuttal 

\cameraready 

\pdfoutput=1
\documentclass[10pt,twocolumn,letterpaper]{article}
\ifreview \usepackage[review]{cvpr} \fi
\ifarxiv \usepackage[pagenumbers]{cvpr} \fi
\ifrebuttal \usepackage[rebuttal]{cvpr} \fi
\ifcamera \usepackage{cvpr} \fi

\usepackage{graphicx}	
\usepackage{amsmath}	
\usepackage{amssymb}	
\usepackage{booktabs}
\usepackage{times}
\usepackage{microtype}
\usepackage{epsfig}
\usepackage[table,xcdraw,dvipsnames]{xcolor}
\usepackage{caption}
\usepackage{float}
\usepackage{placeins}
\usepackage{color, colortbl}
\usepackage{stfloats}
\usepackage{enumitem}
\usepackage{tabularx}
\usepackage{xstring}
\usepackage{multirow}
\usepackage{xspace}
\usepackage{url}
\usepackage{subcaption}
\usepackage{xcolor}
\usepackage[hang,flushmargin]{footmisc}
\usepackage{cuted}
\usepackage{capt-of}

\ifcamera \usepackage[accsupp]{axessibility} \fi

% ...

\ifarxiv  \fi

\newcommand{\R}[1]{{%
    \textbf{%
        \ifstrequal{#1}{1}{\textcolor{red}{R#1}}{%
        \ifstrequal{#1}{2}{\textcolor{blue}{R#1}}{%
        \ifstrequal{#1}{3}{\textcolor{magenta}{R#1}}{%
        \ifstrequal{#1}{4}{\textcolor{teal}{R#1}}{%
                           \textcolor{cyan}{R#1}%
        }}}}%
    }%
}}

\usepackage{xr-hyper}

\makeatletter
\newcommand*{\addFileDependency}[1]{
  \typeout{(#1)}
  \@addtofilelist{#1}
  \IfFileExists{#1}{}{\typeout{No file #1.}}
}

\makeatother

\definecolor{cvprblue}{rgb}{0.21,0.49,0.74}
\usepackage[pagebackref,breaklinks,colorlinks,citecolor=cvprblue]{hyperref}
\usepackage[capitalize]{cleveref}
\crefname{section}{Sec.}{Secs.}
\crefname{table}{Table}{Tables}
\crefname{figure}{Fig.}{Figs.}

\begin{document}
\title{Customize Your Own Paired Data via Few-shot Way}
\author{\authorBlock}
\maketitle

\begin{strip}
\label{sec:teaser}
	\centering
	\includegraphics[width=\textwidth,height=8.2cm]{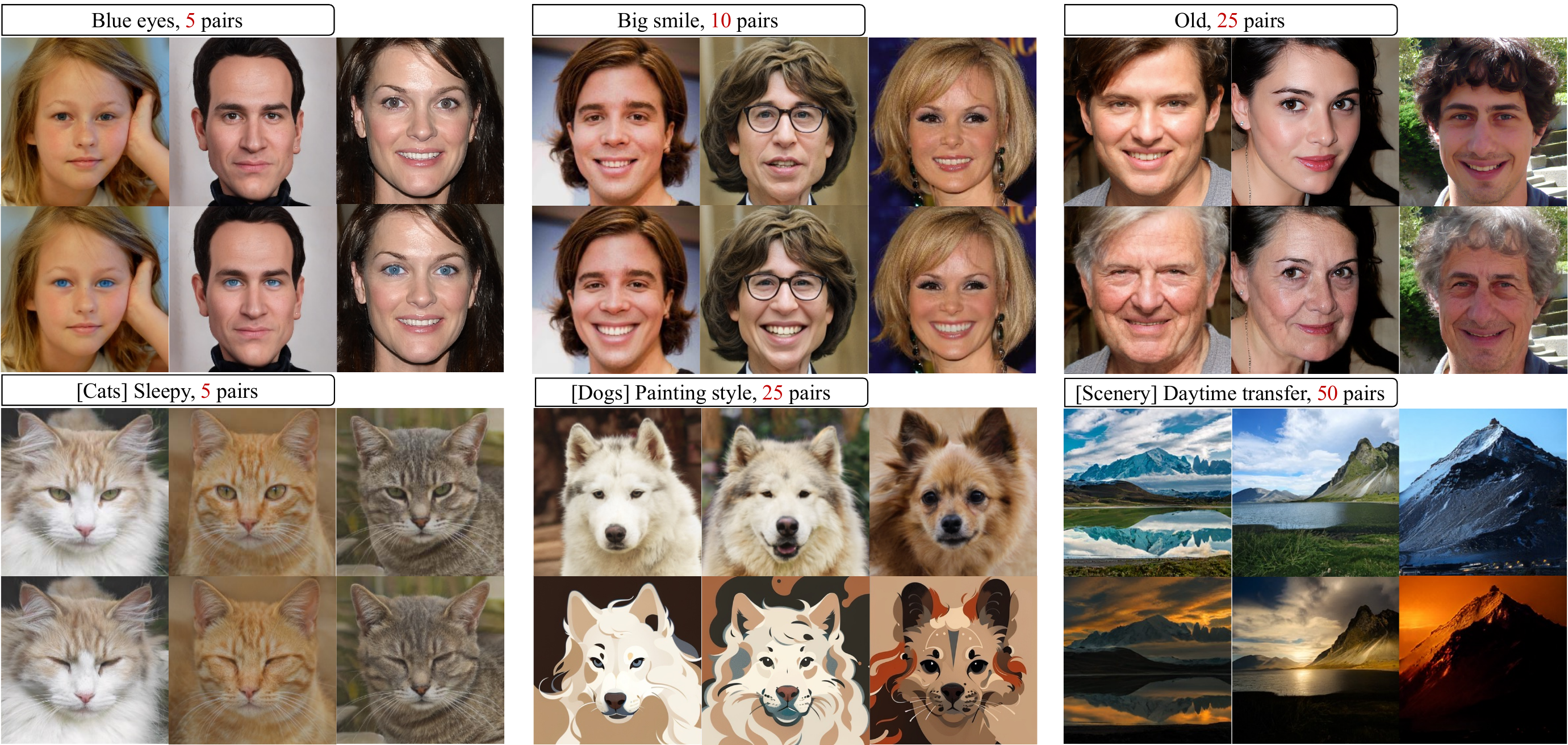}
	\captionof{figure}{Providing only few source-target image pairs, users are allowed to efficiently customize their own image editing models through our framework. Without affecting irrelevant attributes, our proposed method can capture the desired effects precisely. Our model can handle various editing cases, whether the editing targets are commonly known concepts or completely newly defined by users.}
	\label{fig1}
\end{strip}

\begin{abstract}
Existing solutions to image editing tasks suffer from several issues. Though achieving remarkably satisfying generated results, some supervised methods require huge amounts of paired training data, which greatly limits their usages. The other unsupervised methods take full advantage of large-scale pre-trained priors, thus being strictly restricted to the domains where the priors are trained on and behaving badly in out-of-distribution cases. The task we focus on is how to enable the users to customize their desired effects through only few image pairs. In our proposed framework, a novel few-shot learning mechanism based on the directional transformations among samples is introduced and expands the learnable space exponentially. Adopting a diffusion model pipeline, we redesign the condition calculating modules in our model and apply several technical improvements. Experimental results demonstrate the capabilities of our method in various cases.
\end{abstract}
\section{Introduction}
\label{sec:intro}
\hspace{\parindent}Image editing is an important computer vision task which gives the user an interface with the generative models. Based on given instructions, editing models will transfer the original samples into the target domains by modifying certain aspects of the images. Attributes editing and stylizing are all common examples of the image editing tasks. 

Currently, image editing frameworks can be coarsely divided into several categories. As is widely acknowledged, editing tasks can be managed by paired-data-based supervised frameworks like \cite{04_p2p,05_p2phd}, since the original image and the image after edited can be seen as strictly paired data. Subsequently, a few unsupervised editing methods are introduced for the propose of editing images more easily and swiftly. Some proposed unsupervised methods operate directly in the latent space based on certain direction vectors\cite{06_zhu2016generative,43_styleclip,07_huh2020transforming,46_diffae}. Recently, rapid advances in the field of multi-modal generation\cite{01_imagen,02_sd,03_dalle,08_clip,37_karras2019style} have led to new solutions to editing tasks. With the powerful pre-trained priors, such methods allow users to edit with raw sketches, a single reference image or only few words\cite{44_sdedit,45_p2pzero,47_instruct_p2p}.

However, the existing editing methods all have shortcomings to some extent. Firstly, for the methods based on paired data, it is costly to collect a large high-quality paired dataset in most cases. Next, for the methods which edit images in the unsupervised way, bad disentanglement always accompanies with the found implicit directions among different domains, which further leads to unexpected editing effects and makes the editing process imprecise. As for the methods running based on the editing instructions and extra priors, the capabilities of such models are limited by the pre-trained priors as well. For example, frameworks with pre-trained language models such as CLIP \cite{08_clip} cannot handle targets which natural language fails to describe, i.e., users can hardly create new effects beyond the learned domains of provided priors. In a word, there remain difficulties to customize image editing effects expediently, precisely and arbitrarily.

In this work, we get back to the solutions based on paired data, since it is the most direct and reliable way. Given only few pairs of personalized images, we aim to ensure the users to customize their effects efficiently. We first introduce a novel learning mechanism to conquer the overfitting obstacles from which most models suffer when facing small datasets. Instead of training models upon image pairs, we train models upon the directional transformations among samples, thus we can expand the learnable space exponentially. Adopting a diffusion model pipeline, we bring in the pixel-level transformations as the conditions, and redesign the condition injection modules along with some technical improvements to achieve better generation qualities.

As \cref{fig1} and \cref{fig4} show, sufficient experimental results demonstrate the abilities of our proposed method. Achieving an equivalent performance, the amount of the requiring training data of our method is only about 1{\%} of that required by existing paired-data-based methods. There is no disentanglement issue in our method, which means our model focuses on expected editing targets and overlooks other irrelevant attributes. Since running without any task-specific priors, the capability of our model is not limited whether the target effects are commonly known or brand-new concepts created by the users. To summarize, our contributions are as follows:

\begin{itemize}[leftmargin=*]
\item We propose a novel image editing framework, which is driven by only few pairs of data but leads to high-quality generated results. Our approach provides a convenient way for users to customize their own image editing effects.
\item We introduce an efficient few-shot learning mechanism. By making the model learn upon intra-domain transformations rather than cross-domain samples, we expand the learnable space exponentially approximately, which greatly reduces the reliance on large training datasets.
\item Based on a diffusion pipeline, we extend the forms of condition to pixel-space transformations and redesign the whole condition calculating module. Besides, we bring in several slight but effective technical improvements like adaptive noises, skip connections and frequency constraints to our model to achieve better generation qualities.
\item We conduct various experiments to demonstrate the capacities of our framework. Whether it is an attribute editing task or a stylizing task, an entirely newly created effect or a usual effect, an editing task on faces or none-face examples, our model succeeds in completing image editing reliably. Experiments also show that our method outperforms its competitive peers.
\end{itemize}

\section{Related Work}
\label{sec:related}

\subsection{Generative models}

\hspace{\parindent}Generative models continuously play an important role in various computer vision tasks. For a long time, GANs \cite{09_gan,10_denton2015deep,11_radford2015unsupervised,12_salimans2016improved,13_zhao2016energy} have been the main paradigm in the field of image generation, which are able to effectively solve a wide range of important tasks like image in-painting \cite{14_zheng2019pluralistic,15_zhao2020uctgan,16_yi2020contextual}, image-to-image translation \cite{17_anokhin2020high,18_zhang2020cross,19_bhattacharjee2020dunit} and image synthesis \cite{20_shocher2020semantic,21_choi2020stargan,22_lee2020maskgan}. In recent years, diffusion models \cite{23_sohl2015deep,24_ho2020denoising,25_song2020score,26_chen2020wavegrad,27_kingma2021variational,28_kong2020diffwave,29_mittal2021symbolic,30_dhariwal2021diffusion,31_ho2022cascaded,32_saharia2022image} have been proposed as the new SOTA in terms of the generation quality. Adopting a stepwise noising-denoising design, the diffusion models have far finer generation qualities than other self-supervised methods especially on datasets with huge intra-domain variance \cite{33_imagenet,34_schuhmann2021laion}. In this paper, based on a diffusion pipeline \cite{02_sd}, we redesign the forms of conditions and the whole condition-calculating modules. To achieve high-quality image generation, we apply several simple but efficient improvements upon the basic framework.

\subsection{Image editing}
\label{2.2}

\begin{figure*}
\begin{center}
\includegraphics[width=.8\textwidth,height=8cm]{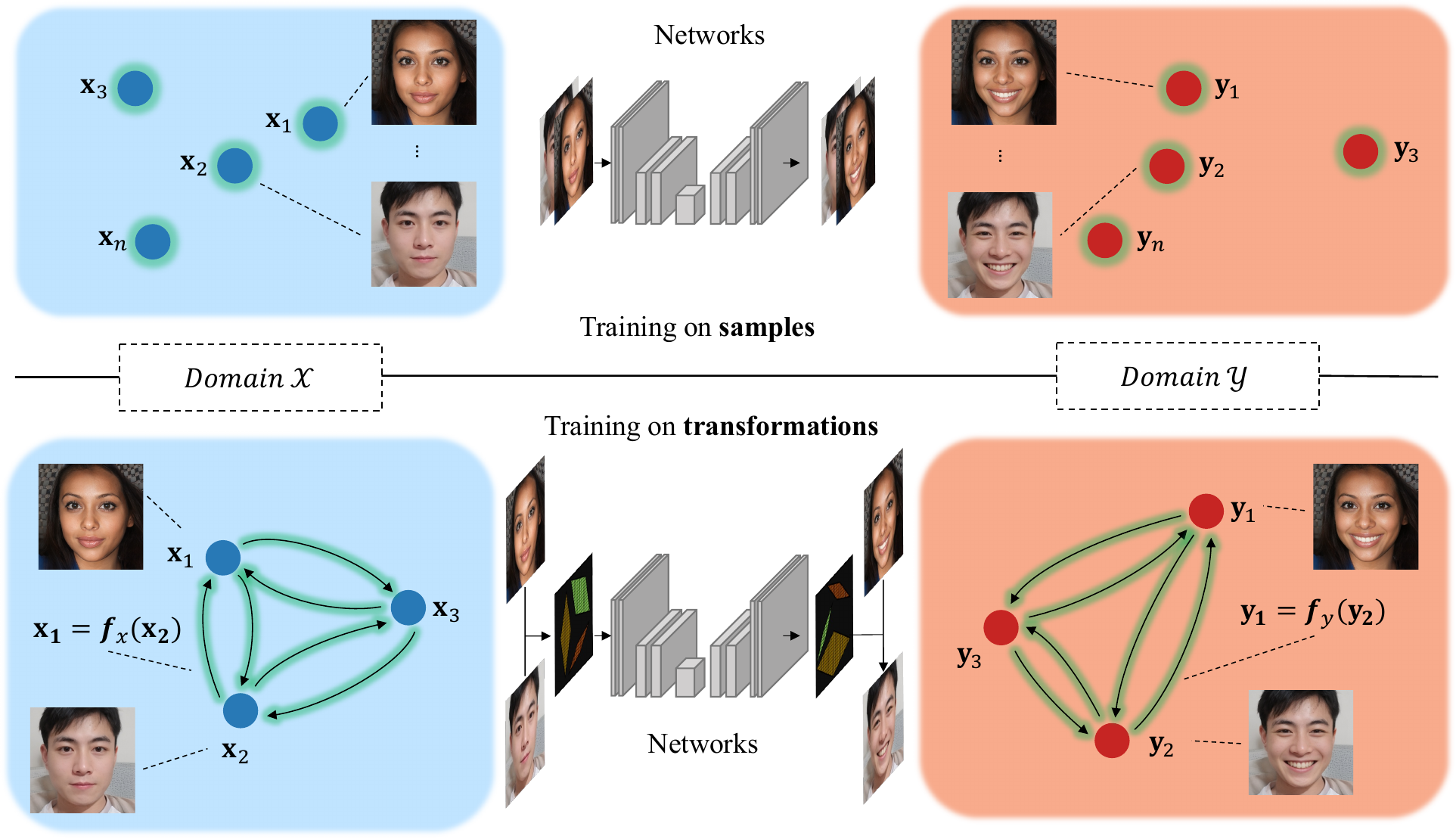}
\end{center}
   \caption{Instead of training the model on the paired samples themselves, we train the model on the directed transformations among samples from the same domain. In the figure we highlight the training objects in green to show the differences between ours and the existing method. In this way we expand the learnable space to an exponential extent approximately.}
   \label{fig2}
\end{figure*}

\hspace{\parindent}Various methods were proposed to solve the image editing tasks. There were a number of approaches \cite{04_p2p,05_p2phd,35_sangkloy2017scribbler,36_spade} that used the paired-data-based frameworks to solve the image editing tasks in early years. The introduction of StyleGAN \cite{37_karras2019style} encouraged a number of distribution-based editing frameworks to be proposed \cite{38_collins2020editing,39_shen2020interpreting,40_harkonen2020ganspace,41_tewari2020stylerig,42_wu2021stylespace,43_styleclip,73_bai2022high}. Such approaches calculate directed vectors between the source and target domains, then operate directly in the latent space, the modulation space or the parameter space based on the found directions. Recently, editing images with multi-modal information and pre-trained large models \cite{44_sdedit,45_p2pzero,46_diffae,47_instruct_p2p,48_prompt2prompt} has been widely accepted by users, and the editing instructions are no longer restricted to the form of images. 

In this work, we return to editing images with paired-data, aiming to accomplish the task through a more cost-effective approach. We consider using paired data directly, which avoids the disentangling problems that may appear in the latent space or in the parameter space of the model. Besides, paired-data-based methods are far more suitable for customized image editing cases than methods using pre-trained models, as pre-trained models cannot manage tasks beyond their training domains. To conquer the heavy reliance on large paired dataset, we propose to expand the dataset itself rather than introducing any additional priors, which would avoid the model being limited by the capabilities imposed by the priors themselves.

\section{Method}
\label{sec:method}

\subsection{Expansion methods for few-shot datasets}
\label{sec3.1}

\hspace{\parindent}Given the source domain $\mathcal X$ and the target domain $\mathcal Y$, the editing model marked as $M$ will be trained on pairs marked as $[\mathbf x,\mathbf y]$ where $\mathbf x\in\mathcal X$ and $\mathbf y\in\mathcal Y$. We can simply conclude the image editing task as training the model $M$ to fit the function $\mathbf y=M(\mathbf x)$ . We call such method of training on $[\mathbf x,\mathbf y]$ pairs ``one-source-to-one-target" for short. Mark $\mathit m$ as the total number of the training pairs, such training process works well when $\mathit m$ is large. However, when it comes to the few-shot cases, $M$ tends to overfitting the mapping relationships provided by little training data, because the learnable space of $M$ is correlated with  $\mathit m$ only to a linear extent. Phenomenons like mode collapse indicate that when the generative model overfits, it has learned only few patterns and fails on samples outside the training domain.

We introduce a novel approach which expands the learnable space of the training dataset without bringing in any other class-related priors. Instead of training $M$ on batches of paired data, we consider to make $M$ train upon the transform function $\mathit f$, which conducts transformations {\bf inside} the batches of paired data. Specifically, we can randomly sample $\mathit n$ pairs of training data, marked as \{$[\mathbf x_1,\mathbf y_1], [\mathbf x_2,\mathbf y_2], ..., [\mathbf x_n,\mathbf y_n]$\}, from the total training dataset with replacement. Among samples inside such group, obviously, there must be functions $\mathit f_x$ and $\mathit f_y$ which complete the one-to-the-other transformations, i.e., for $\forall \mathit i,\ 1\leq \mathit i \leq \mathit n$, $\mathit i \in \mathbb {N}^+$, $\exists\ (\mathit f_x, \mathit f_y)$ satisfies $\mathbf x_i = \mathit f_x (\{\mathbf x_j\})$ and $\mathbf y_i = \mathit f_y (\{\mathbf y_j\})$, where $1\leq \mathit j \leq \mathit n, j \neq i, \mathit j \in \mathbb {N}^+$.

Here we start off with the simplest case where $\mathit n=2$. To construct each \{$[\mathbf x_1,\mathbf y_1], [\mathbf x_2,\mathbf y_2]$\}, given $[\mathbf x_1,\mathbf y_1]$, there will be $C_m^1$ additional $[\mathbf x_2,\mathbf y_2]$ to be selected from the training dataset. Namely, if we set the training target of $\mathit M$ to fitting $\mathit \mathit f_y=\mathit M (\mathit f_x)$, there are $\mathit {m}*C_m^1$ pairs of $[\mathit f_x, \mathit f_y]$ for the model to learn from, which are far larger than the number of $[\mathbf x,\mathbf y]$, i.e., $\mathit m$. \cref{fig2} is an illustration of $\mathit n=2$ case. To distinguish from the ``one-source-to-one-target" method, we call our method ``$\mathit n$-source-to-$\mathit n$-target". $\mathit n$ can be assigned to values larger than 2, and a larger $\mathit n$ means the expansion's degree of the training dataset is larger. The degree of expansion refers to $\mathit {m}*C_m^{n-1}$, which approximately leads to an exponential growth of the learnable space. We adopt the setting of $\mathit n=2$ in all following statements and experiments, while a further discussion about other settings of $\mathit n$ value is in the appendix.

{\bf Reasonability:} Consider the simplest case: suppose that $\exists \mathbf x_1, \mathbf x_2$, $\mathbf x_2$ is the version of $\mathbf x_1$ after horizontally flipped, then the corresponding $\mathbf y_2$ should be the version of $\mathbf y_1$ after horizontally flipped as well. In this simple case, both $\mathit f_x$ and $\mathit f_y$ equal the function of horizontal flip, namely $\mathit f_x = \mathit f_y$. Similar conclusions can be drawn when $\mathit f$ is simple spatial transform like vertical flip, rotation, padding and so on. In more general cases, the differences between two images often represent as different id information, different poses, etc., where $\mathit f_x$ is unlikely to straightly equal to $\mathit f_y$. However, perceptually we find that complex transformations can always be decomposed into linear or non-linear combinations of fundamental transformation functions. Hence, there is certain manifold upon which the projections of $\mathit f_x$ and $\mathit f_y$ are equal. An illustration \cref{fig7} is shown for the better understanding. The explanation mentioned above suggests that training $\mathit M$ on $(\mathit f_x, \mathit f_y)$ pairs is reasonable and able to converge.

\begin{figure}[tp]
\begin{center}
\includegraphics[width=\linewidth]{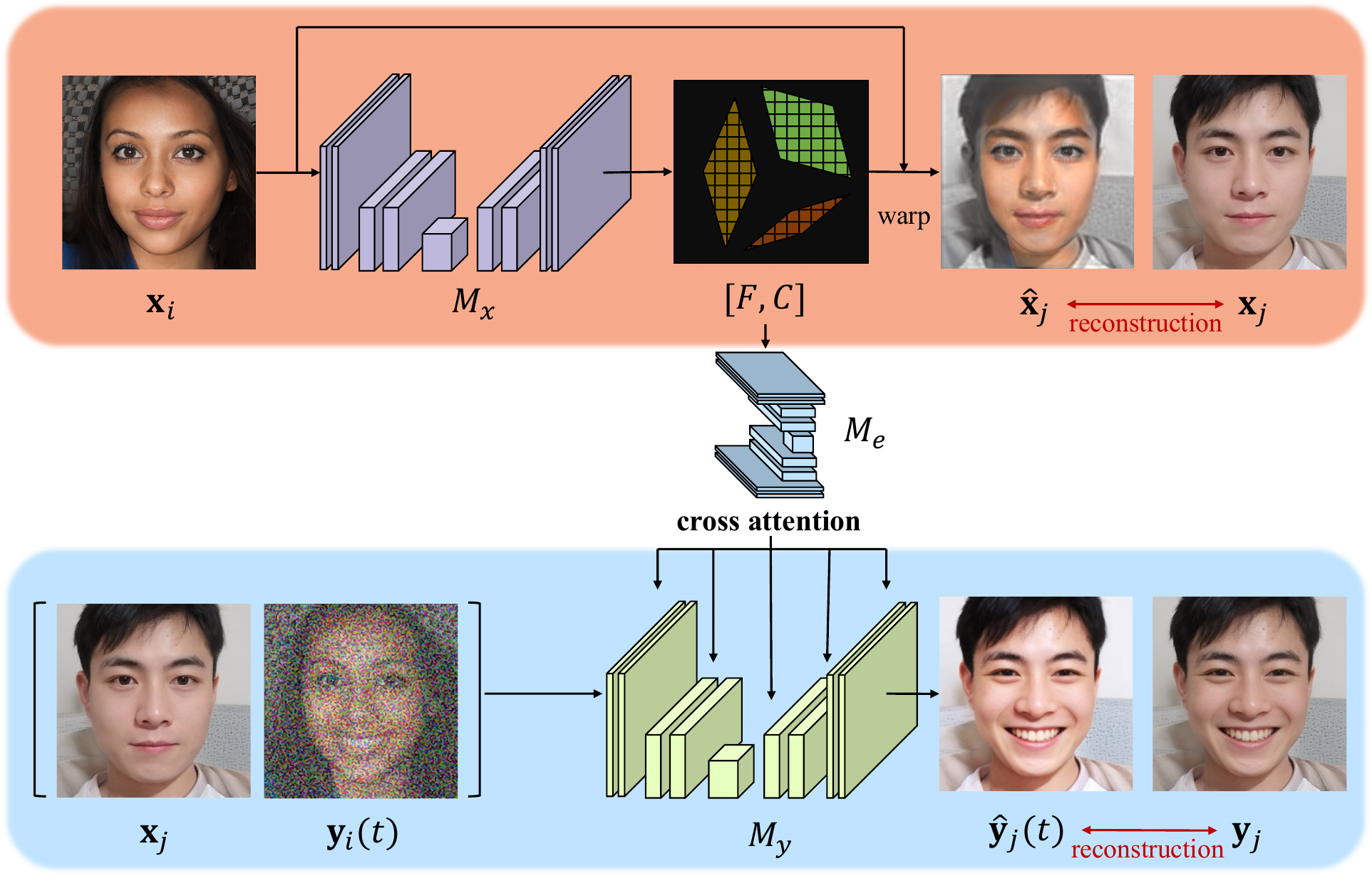}
\end{center}
   \caption{All modules of our framework are trained jointly and end-to-end. During training, all samples mentioned in the illustration are selected from the whole training dataset, while $\mathbf x_j$ is provided by users and will be transferred to get $\mathbf y_j$ for the inference time.}
\label{fig3}
\end{figure}

\subsection{Approach}

\hspace{\parindent}We build two models, marked as $\mathit M_x$ and $\mathit M_y$ respectively, to calculate the transform relationships $\mathit f_x$ and $\mathit f_y$. The overall model $\mathit M$ runs to convert $\mathit f_x$ to $\mathit f_y$. More precisely, we make $\mathit M_x$ transform $\mathbf x_i$ to $\mathbf x_j$, thus we obtain $\mathit f_x$. Due to the relationships between $\mathit f_x$ and $\mathit f_y$ explained in \cref{sec3.1}, $\mathit f_x$ can assist $\mathit M_y$ to transform $\mathbf y_i$ to $\mathbf y_j$, where $\mathbf y_j$ is exactly the edited target we need. All these models are trained jointly and the whole training process is end-to-end. For the training procedure, $\mathbf x_i$, $\mathbf x_j$, $\mathbf y_i$ and $\mathbf y_j$ are all randomly sampled from the training datasets. As for the test time, $\mathbf x_i$, $\mathbf y_i$ are still random training samples, while $\mathbf x_j$ refers to the image need to be edited and $\mathbf y_j$ is the expected result. The complete framework is shown in \cref{fig3}.

In terms of the conversion among samples of the source domain $\mathcal X$, we simplify the transformation between two images to spatial transformations $\mathcal T_s$ combined with color shifting $\mathcal T_c$, both of which are kinds of explicit representations. Inspired by existing methods \cite{54_fomm}, optical flow per pixel is chosen to embody spatial transforms $\mathcal T_s$. Mark a flow-field map as $\mathit F \in \mathbb {R}^{H \times W \times 2}$, where $\mathit H, \mathit W$ equals the height and width of the image respectively. With the directions of the height $ \mathit H $ and width $ \mathit W $ used as the coordinate axes, $\forall \mathit h, \mathit w\in \mathbb {N}^+$, $1\leq \mathit h \leq \mathit H$, $1\leq \mathit w \leq \mathit W$, the value at the position of $(\mathit h, \mathit w)$, i.e., $\mathit F_{\mathit h,\mathit w} \in \mathbb {R}^2$, represents a set of positions in the reference image. Given a reference image $\mathit I_{ref}$, mark the warped image after applying $\mathit F$ as $\mathit I_{flow}$. To calculate the value at each position of $\mathit I_{flow}$, get the value at the same position of $\mathit F$, treat such value of two dimensions as a set of coordinates, and assign the value from $\mathit I_{ref}$ which locates in such coordinates to pixels of $\mathit I_{flow}$. \cref{eq2} expresses the spatial transformation $\mathcal T_s$ mentioned above.

\begin{equation}
\label{eq2}
\begin{aligned}
\mathit I^{flow}_{\mathit h,\mathit w}=\mathcal T_s (I^{ref})=\mathit I^{ref}_{F_{(h,{w)_1}},F_{(h,{w)_2}}}
\end{aligned}
\end{equation}

As for color shifting $\mathcal T_c$, we use 2D linear affine transformations marked as $\mathit C \in \mathbb{R}^{\mathit H\times\mathit W\times2}$, where $\mathit C_{\mathit h, \mathit w} \in \mathit{R}^2$ denotes a set of affine parameters. In particular, given a reference image $\mathit I_{ref}$, mark the version of $\mathit I_{ref}$ after affine transformed as $\mathit I_{aff}$, then the color shifting $\mathcal T_c$ can be calculated as \cref{eq3}.

\begin{equation}
\label{eq3}
\begin{aligned}
\mathit I^{aff}_{\mathit h,\mathit w}=\mathcal T_c (I^{ref})=C_{(\mathit h, \mathit w)_1} * \mathit I^{ref}_{\mathit h,\mathit w} + C_{(\mathit h, \mathit w)_2}
\end{aligned}
\end{equation}

According to the above explanation, $\mathit M_x$ takes in $\mathbf x_i$ and outputs a set of transformations containing $\mathit F$ and $\mathit C$. Applying $\mathit F$ and $\mathit C$ to $\mathbf x_i$ in sequence, we obtain the predicted result $\mathbf {\hat x}_j$, which is described in \cref{eq4}.

\begin{equation}
\label{eq4}
\begin{aligned}
\mathbf {\hat x}_j = \mathcal T_c(\mathcal T_s(\mathbf x_i))
\end{aligned}
\end{equation}

By setting pixel-level reconstruction loss like MSE between $\mathbf {\hat x}_j$ and $\mathbf x_j$, $\mathit M_x$ is trained to finish the image editing task in the source domain, whose output marked as $[F,C] \in \mathbb R^{H\times W\times4}$ represents the explicit transformations between $\mathbf {\hat x}_j$ and $\mathbf x_j$, i.e., $\mathit f_x$.

\begin{figure}[tp]
\begin{center}
\includegraphics[width=\linewidth]{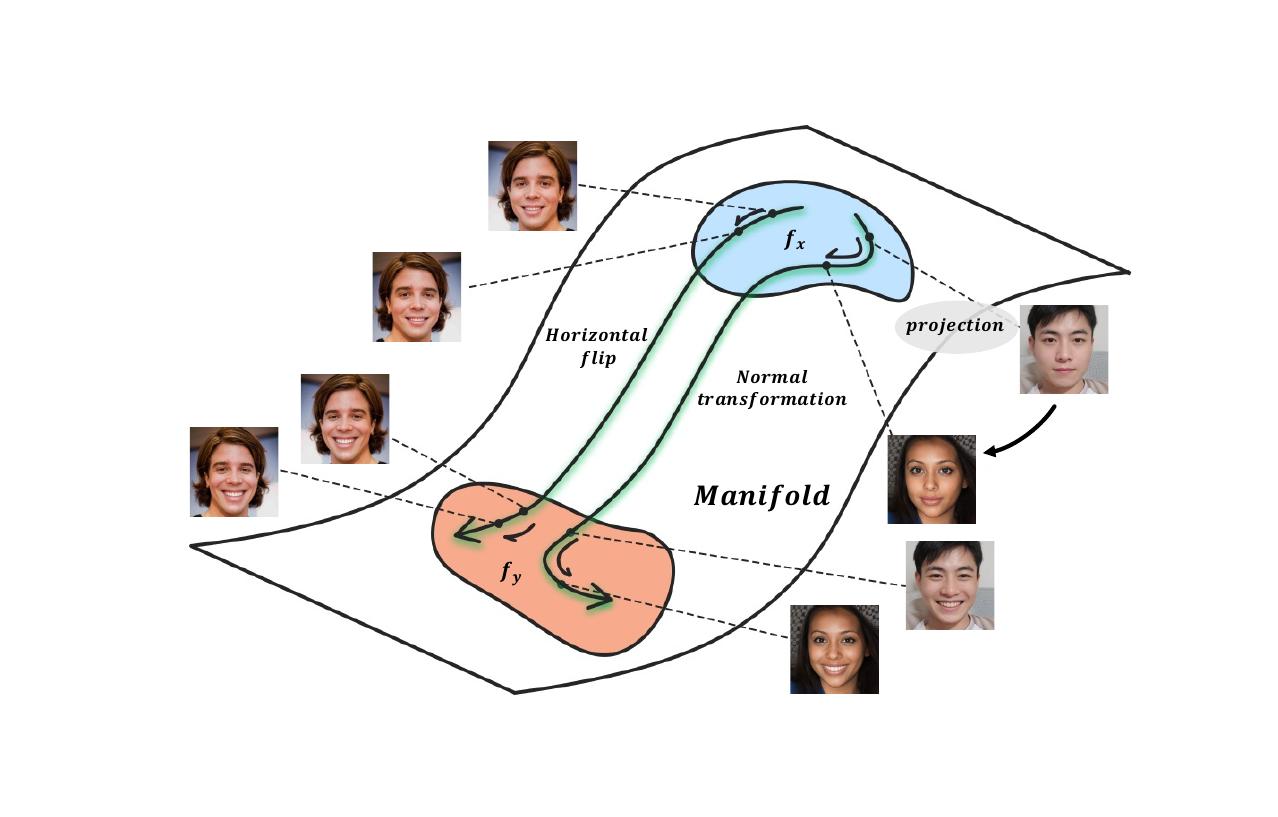}
\end{center}
   \caption{Essentially, our model is trained to find certain manifold where the projections of $\mathit f_x$ and $\mathit f_y$ are equal. Horizontal flip is shown on the upper left as a simple example, and normal cases are shown on the lower right.}
\label{fig7}
\end{figure}

Compared with the conversion among images of the source domain $\mathcal X$, the translation in the target domain $\mathcal Y$ is what we really care about in editing tasks. To ensure the high qualities of the edited results $\mathbf {y}_j$, we adopt a diffusion pipeline to train $\mathit M_y$. The way of adding noises of the pipeline obeys a regular linear approach. The value of $\mathit t$ varies during training. Hypermeters related to $\mathit t$ determine the degree of noise added to the image. A noise marked as $\mathbf z$ is randomly sampled from the Gaussian distribution $\mathcal N(0,I)$. The final time-specific results combined by $\mathbf {y}_i$ and $\mathbf z$ is marked as $\mathbf y_i(t)$. Furthermore, we concatenate $\mathbf x_j$ to $\mathbf y_i(t)$ and treat such $\mathbf y_i(t) | \mathbf x_j$ as the input of $\mathit M_y$, which makes the whole training process of the framework more stable.

A difference between ours and other methods is the training target of $\mathit M_y$ is to reconstruct neither $\mathbf y_i(0)$ nor $\mathbf z$ but $\mathbf y_j$. $[F,C]$ is considered as condition information besides $\mathbf x_j$, which will act on $\mathit M_y$ through cross-attention modules after passing an additional embedding module marked as $\mathit M_e$. A discussion about this algorithm part is in our appendix.

\subsection{Technical details and improvements}

\hspace{\parindent}Our overall framework is inspired by \cite{02_sd}: $\mathit M_y$ runs between a pair of encoder-decoder, which makes the framework work in the semantic space. There are two kinds of conditions in our framework, one is $\mathbf x_i$ which works by concatenation, and the other is $\mathit {M}_e([F,C])$ coming from $\mathit M_x$ which works through cross-attention modules.

We notice that some methods use frozen pre-trained encoder-decoder, i.e., a variational auto-encoder or VAE \cite{55_vae} namely, in their diffusion models when training from the scratch or fine-tuning on their customized tasks. Such VAEs are well-trained on huge datasets along with traditional distribution-based and reconstruction losses. However, VAEs are always criticized for blurred and oversmoothing generated images. Then some existing methods conducting an extra online super-resolution behind the output of the decoder as a remedy. In our approach, we take the following steps to achieve high-quality image generation directly: Train the VAE together with the other parts of the framework without freezing any parameters, add adaptive noises to the intermediate layers of the decoder and set skip connections between relative layers of the encoder and the decoder.

Since small datasets have a far stronger bias than huge datasets, the knowledge learned from huge datasets\cite{33_imagenet} may be inappropriate for training models in customized few-shot cases. We treat the VAE as a trainable sub-module, which acts like not ``semantic" compressing but only ``dimension" compressing, then the VAE is more likely to adjust to the specific cases provided by the users. Adding noises and skip connections are common approaches in existing methods \cite{37_karras2019style}. We observe that these two tricks enable the model to capture the high frequency information, thus increasing the details of the generated samples to a large extent.

Additionally, following Dits \cite{56_dits}, we replace the backbone U-Net of $\mathit M_y$ with ViTs \cite{57_vits} and add dual layer normalization layers \cite{58_dual}, by which we observe an improvement on the quality of generated samples. To further enhance the generalization of the model, we apply augmentation methods to $\mathbf x_j$ and $\mathbf y_j$. Augmentations we use include horizontal flip, color jitter and random affine, which are all common in existing frameworks. Moreover, a constraint in the frequency domain \cite{62_freq} is applied which makes the model coverage faster and perform better especially in details.

\begin{figure*}
\begin{center}
\includegraphics[width=0.93\textwidth,height=18.5cm]{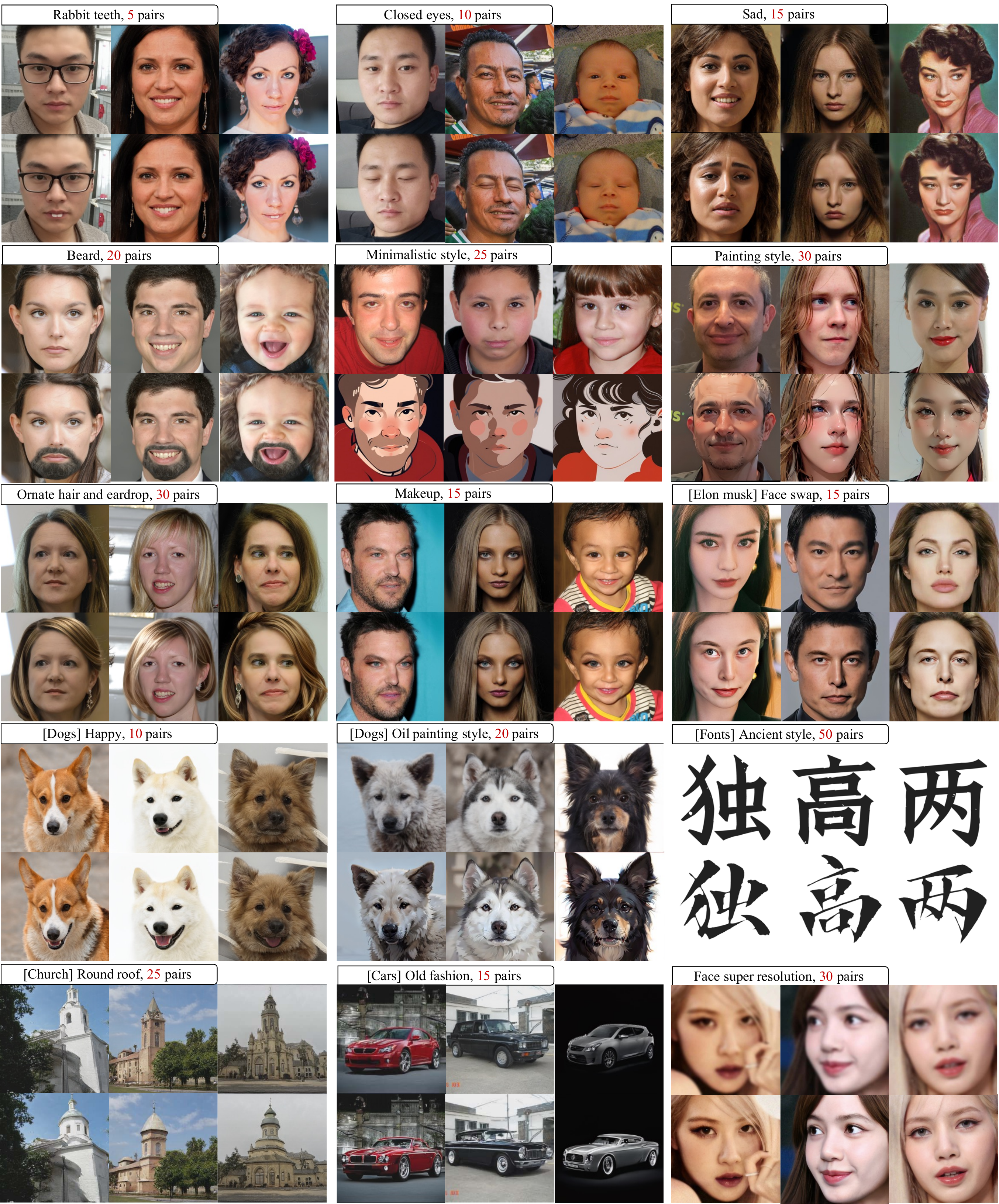}
\end{center}
   \caption{Here we show more experimental results of our method. The editing target and the amount of given training pairs are recorded on the top of each result unit, and the relative generated images are displayed below. More visual results are shown in our appendix.}
   \label{fig4}
\end{figure*}
\section{Experiments}
\label{sec:Experiments}

\begin{figure*}
\begin{center}
\includegraphics[width=0.95\textwidth,height=9.5cm]{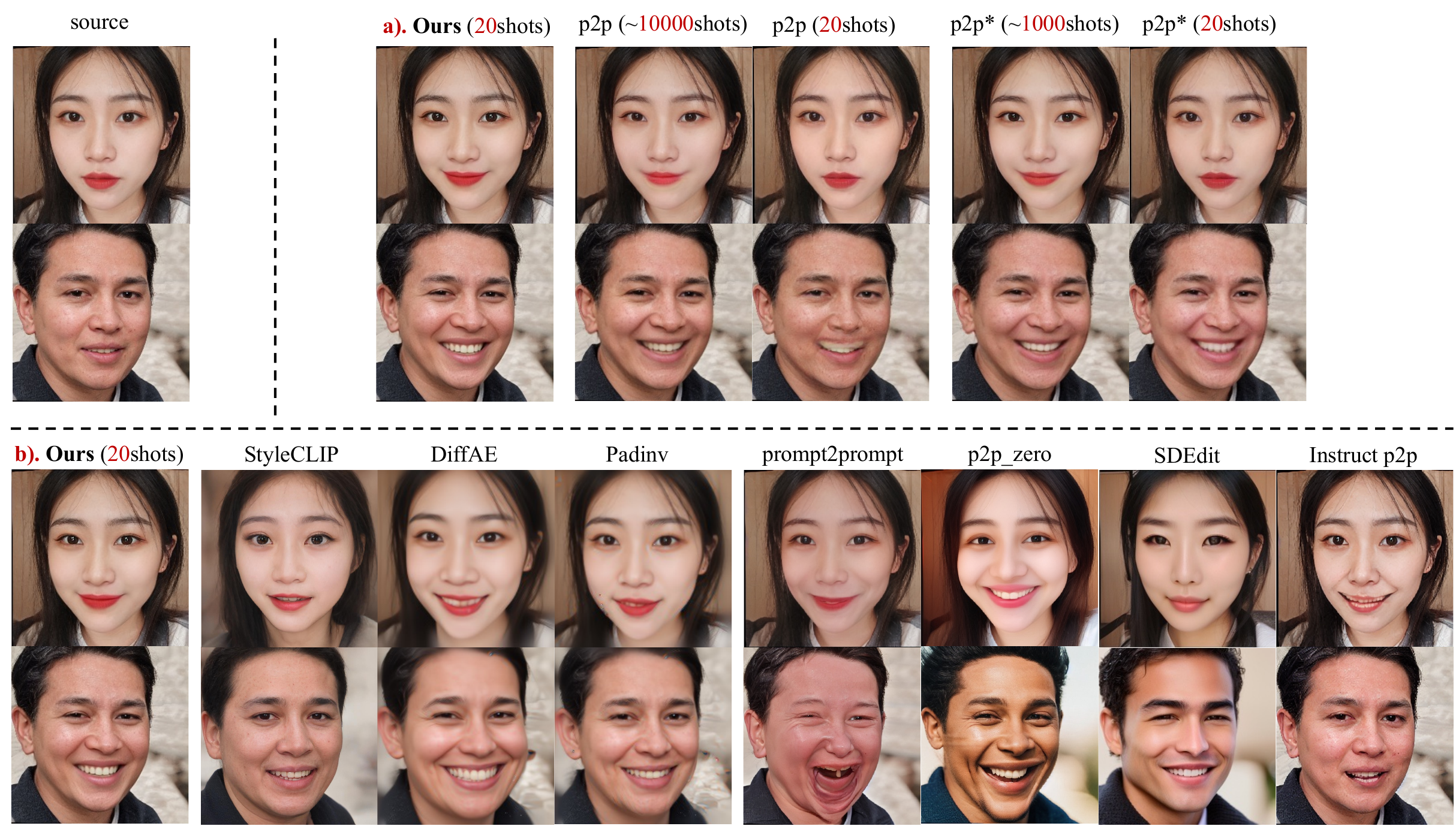}
\end{center}
   \caption{Comparisons between existing works and our method. Part \textbf{a)} shows comparisons with paired-data methods, while part \textbf{b)}  contains results comparing with few/zero-shot methods including methods editing in the latent space and methods relying on the multi-modal information. Note that for the latter methods, we will provide proper instructions which can describe the editing target.}
\label{fig5}
\end{figure*}

\subsection{Generated results}

\hspace{\parindent}Generated results of our framework are shown in \cref{fig1} and \cref{fig4}. The editing targets and the amount of given training data pairs are recorded on the top of each result unit in the figure, and the relative generated samples are shown below. Our experiments are conducted on both 256px and 512px. To thoroughly test the capability of the model, experiments are carried out on a wide field of topics. More visual results, training hyper-parameters, detailed experimental settings, experiments about proper $\mathit m$ value for different topics, and a discussion about the limitations of our method are shown in our appendix.

\subsection{Comparison}

\hspace{\parindent}We roughly divide the competitive peers of our method into two types: paired data based methods and unpaired data based methods, according to whether using paired training data. For comparing with methods that training with paired data, we primarily care about the trade-off between the generated qualities and the number of training samples. For comparing with methods that are independent of paired data, we will show different performances of mentioned methods with comparable data costs. For all the baselines, we use the source codes and parameter settings released officially.

{\bf Evaluation settings.} For each experiment of the fair comparison, we firstly run all the methods through different amounts of training data in order to find out how much paired data is needed for each method to behave best. Besides recording the performances got on an adequate amount, we record the results of the competitors when they run with the same amount of data as ours for a more clear comparison. For the unfair comparison, we provide proper samples according to the actual needs of the methods, and we test the performances of our framework with comparable data costs.

{\bf Metrics.} There are several criteria used when comparing performances. Firstly, FID \cite{69_fid} is calculated to measure the similarity between the generated samples and the real ones. Furthermore, to measure the degree of clarity, PSNR is calculated. Moreover, we introduce another metric named DIoU in the case of available paired data. Given a pair of samples, we can obtain the editing targets by subtracting the source image from its paired target image directly in the pixel space. Finding the unions of all editing target areas of the training data, we can assign it as the expected targets of the total task. In a similar way we can calculate the actual edited areas generated by the model. Then the intersection over those two kinds of unions, namely IoU, may represent the ``accuracy" of the model when editing images. A lower DIoU, the IoU of Diversities generated by the model, means the model learns the editing targets more precisely.

{\bf Main conclusions.} The qualitative results are shown in \cref{fig5} and quantitative results are in \cref{tab1}. As a more general evaluation, the user study is shown in our appendix.

For the part of fair comparisons, we train pix2pix \cite{04_p2p}, pix2pix* and our method upon a number of different editing cases. Note that pix2pix* is an improved version of pix2pix with a few technical tricks whose detailed settings can be found in the appendix. Obviously, our method can achieve an equivalent performance compared to its competitive peers with only 10{\%} or even 1{\%} of the training samples. Such amounts of samples are absolutely affordable when users hope to customize their own image editing tasks. A further comparison about the training time is in the appendix.

For the experiments of unfair comparisons, we select a series of methods to compare with including DiffAE \cite{46_diffae}, Styleclip \cite{43_styleclip}, Padinv \cite{73_bai2022high}, prompt2prompt \cite{48_prompt2prompt}, pix2pix-zero \cite{45_p2pzero}, SDEdit \cite{44_sdedit} and instruct-pix2pix \cite{47_instruct_p2p}. We choose these editing methods to compare with because they all run with little data cost. However, it is worth emphasizing that these methods use strong priors like large well-trained models and are independent of paired samples, so they are not strictly competitive peers of our method. Actually for such methods editing in the domain level, disentanglement has been widely acknowledged as a troublesome problem. If the editing tasks require fine-grained operations, these methods tend to producing unexpected changes beyond the editing areas. Moreover, these frameworks can only fetch and combine {\bf known} concepts from the priors' training spaces, which means users can hardly {\bf create} their own effects from scratch. As \cref{fig5} shows, visually, our method manages to handle few-shot cases. Furthermore, areas outside the editing targets preserve quite well in our generated samples. As our framework runs without any priors, there are no restrictions on the capacities of the model.

\begin{table}\small
\begin{center}
\begin{tabular}{cccc}
\hline
\bf{Methods} & \bf{FID}$\downarrow$ & \bf{PSNR}$\uparrow$ & \bf{DIoU}$\uparrow$ \\
\hline
Pix2pix-20shots & 0.2971 & 17.9045 & 0.8253 \\
Pix2pix$ \sim $10000shots & 0.0176 & 24.5218 & 0.8755 \\
Pix2pix*-20shots & 0.1882 & 20.0071 & 0.8401 \\
Pix2pix*$ \sim $1000shots & 0.0151 & 26.6419 & 0.9037 \\
\hline
StyleCLIP & 0.0332 & 27.4288 & 0.6506 \\
DiffAE & 0.2580 & 27.5360 & 0.7476 \\
Padinv & 0.0721 & 27.1953 & 0.7380 \\
prompt2prompt & 0.4725 & 27.0361 & 0.5628 \\
pix2pix-zero & 0.3617 & 27.8343 & 0.5054 \\
SDEdit & 0.1764 & 27.4527 & 0.6822 \\
Instruct-pix2pix & 0.1490 & 27.1438 & 0.7779 \\
\hline
\textbf{Ours}-20shots & \bf0.0094 & \bf27.8455 & \bf0.9352 \\
\hline
\end{tabular}
\end{center}
\caption{The quantitative results between ours and existing methods. The metrics are calculated on 1000 randomly generated samples. All values recorded in the table are averages across multi-repeating experiments.}
\label{tab1}
\end{table}
\section{Ablation study}
\label{sec:ablation}

\subsection {Effects of the data pairing mechanism}

\begin{figure}[tp]
\begin{center}
\includegraphics[width=\linewidth]{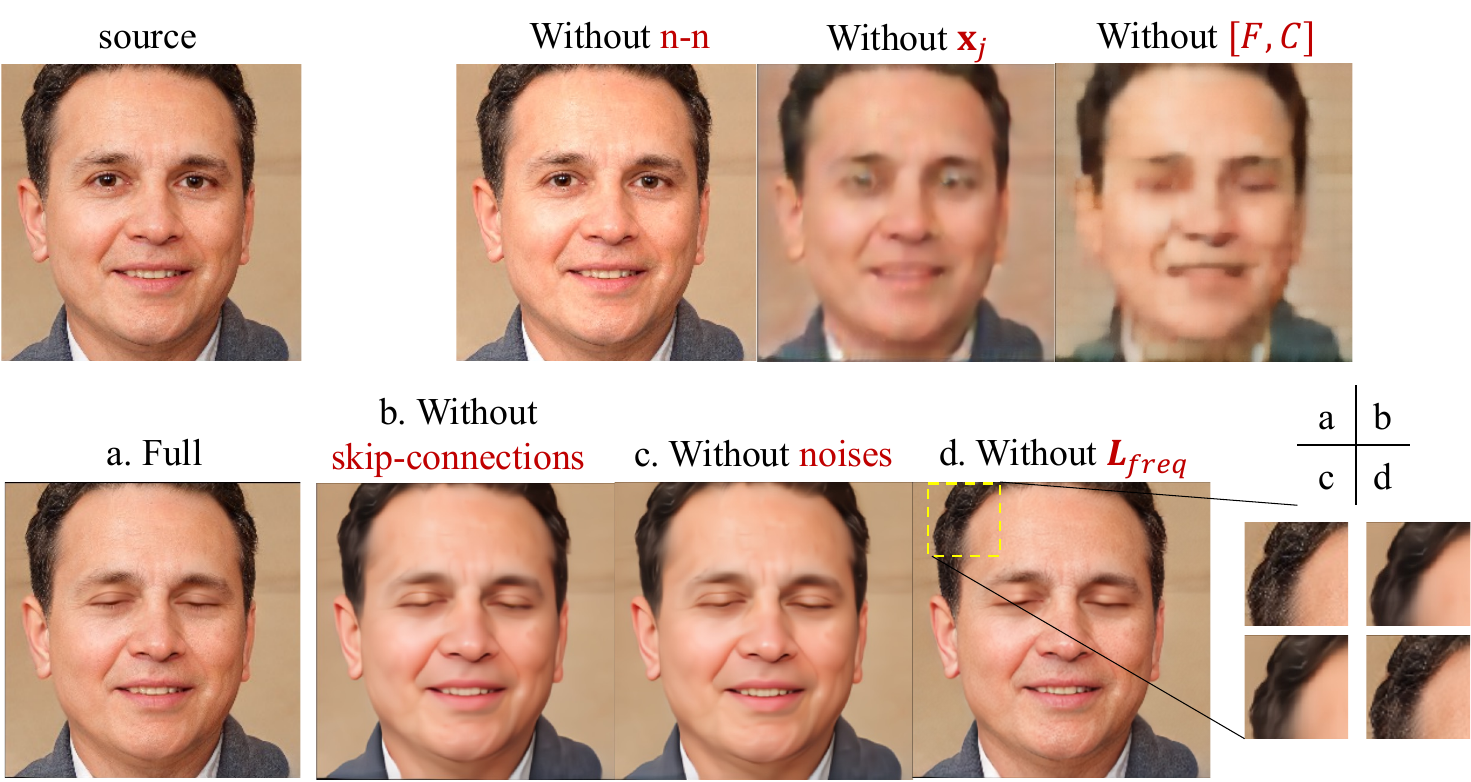}
\end{center}
   \caption{The qualitative results of the ablation study. Without our ``$\mathit n$-source-to-$\mathit n$-target" method, the model overfits easily. Both $\mathbf x_j$ and $[F,C]$ ease the difficulties of the training. The improvements we adopt help the model to generate more fine-grained results.}
\label{fig6}
\end{figure}

\hspace{\parindent}As \cref {sec3.1} explains, the expansion method is the key to make the whole framework run with few samples. We set a pair of experiments, one of which is the original model we proposed, and the other returns to a normal ``one-source-to-one-target" methods like ways used in existing models, i.e., both $\mathit M_x$ and $\mathit M_y$ take in $\mathbf x_i$ and output $\mathbf y_i$. Results in \cref{fig6} and \cref{tabA6_ablation_study} show that the ``one-source-to-one-target" method gets stuck in overfitting easily with few training samples, while our ``$\mathit n$-source-to-$\mathit n$-target" method runs stably.

\subsection{Effects of the condition}

\hspace{\parindent}There are two kinds of condition in our framework: $\mathbf x_j$ concatenated to $\mathbf y_i(t)$, and $[F,C]$ output by $\mathit M_x$. The first experiment is conducted to observe the influence if $\mathbf x_j$ is not provided as a condition. As \cref{fig6} shows, $\mathbf x_j$ greatly decreases the difficulties of the training process, making the model converge more easily. The second experiment aims to evaluate the effects of $[F,C]$. We disable $\mathit M_{e}[F,C]$ by setting the input conditions of cross attention module to a constant and train the whole model, which forces the model to train to fit $\mathbf y_j$ only based on $\mathbf y_i(t)$ and $\mathbf x_j$. Without a proper guidance, we observe a similar phenomenon that the model is hard to converge. Especially, it takes a much longer time for $\mathit M_y$ to generate the overall contour shape of $\mathbf y_j$. Finally we draw the conclusion that such two conditions provide a basic ``canvas", thus the model only need to focus on the differences between domain $\mathcal X$ and domain $\mathcal Y$.

\begin{table}\small
\begin{center}
\begin{tabular}{cccc}
\hline
{\bf Setting} & {\bf FID$\downarrow$} & {\bf PSNR$\uparrow$} & {\bf DIoU$\uparrow$} \\
\hline
Baseline &\bf0.0094 & \bf27.8455 & \bf0.9352\\
without ``$\mathit n$ - $\mathit n$" & 0.9635 & 20.6481 & 0.2830\\
without $\mathbf x_{j}$ & 0.5502 & 17.5893 & 0.1894\\
without $[F,C]$ & 0.6938 & 17.2037 & 0.2043\\
without skip-connections & 0.0256 & 25.4526& 0.3451\\
without noises & 0.0127 & 26.4819 & 0.7129 \\
without $L_{freq}$ & 0.0115 & 26.9471& 0.8125 \\
\hline
\end{tabular}
\end{center}
\caption{The quantitative results of the ablation study. Recorded values are consistent with conclusions drawn from \cref{fig6}.}
\label{tabA6_ablation_study}
\end{table}

\subsection{Ablation study of the technical tricks}

\hspace{\parindent}We set a few experiments to figure out the actual effect of each technical tricks. As \cref{fig6} shows, adding skip connections assists the model in generating details, while adding noises and applying frequency loss work similarly to enrich the information in extremely high frequency domains. The subplot on the bottom right of \cref{fig6} shows the different performances of the mentioned methods in generating high frequency details.
\section{Conclusion}
\label{sec:conclusion}

\hspace{\parindent}We propose a novel framework which encourages users to customize their own editing models with only few pairs of training data. Our model is based on an efficient dataset expansion method, a robust diffusion pipeline, a few redesigned modules and technical improvements. Experiments demonstrate that our model is able to handle different topics of editing tasks and has good qualities of generated images. Further discussions about limitations, future work, and other details are shown in the appendix.

{\small
\bibliographystyle{ieeenat_fullname}
\bibliography{11_references}

\begin{thebibliography}{56}
\providecommand{\natexlab}[1]{#1}
\providecommand{\url}[1]{\texttt{#1}}
\expandafter\ifx\csname urlstyle\endcsname\relax
  \providecommand{\doi}[1]{doi: #1}\else
  \providecommand{\doi}{doi: \begingroup \urlstyle{rm}\Url}\fi

\bibitem[Anokhin et~al.(2020)Anokhin, Solovev, Korzhenkov, Kharlamov, Khakhulin, Silvestrov, Nikolenko, Lempitsky, and Sterkin]{17_anokhin2020high}
Ivan Anokhin, Pavel Solovev, Denis Korzhenkov, Alexey Kharlamov, Taras Khakhulin, Aleksei Silvestrov, Sergey Nikolenko, Victor Lempitsky, and Gleb Sterkin.
\newblock High-resolution daytime translation without domain labels.
\newblock In \emph{Proceedings of the IEEE/CVF Conference on Computer Vision and Pattern Recognition}, pages 7488--7497, 2020.

\bibitem[Bai et~al.(2022)Bai, Xu, Zhu, Xia, Yang, and Shen]{73_bai2022high}
Qingyan Bai, Yinghao Xu, Jiapeng Zhu, Weihao Xia, Yujiu Yang, and Yujun Shen.
\newblock High-fidelity gan inversion with padding space.
\newblock In \emph{European Conference on Computer Vision}, pages 36--53. Springer, 2022.

\bibitem[Bhattacharjee et~al.(2020)Bhattacharjee, Kim, Vizier, and Salzmann]{19_bhattacharjee2020dunit}
Deblina Bhattacharjee, Seungryong Kim, Guillaume Vizier, and Mathieu Salzmann.
\newblock Dunit: Detection-based unsupervised image-to-image translation.
\newblock In \emph{Proceedings of the IEEE/CVF Conference on Computer Vision and Pattern Recognition}, pages 4787--4796, 2020.

\bibitem[Brooks et~al.(2022)Brooks, Holynski, and Efros]{47_instruct_p2p}
Tim Brooks, Aleksander Holynski, and Alexei~A Efros.
\newblock Instructpix2pix: Learning to follow image editing instructions.
\newblock \emph{arXiv preprint arXiv:2211.09800}, 2022.

\bibitem[Chen et~al.(2020)Chen, Zhang, Zen, Weiss, Norouzi, and Chan]{26_chen2020wavegrad}
Nanxin Chen, Yu Zhang, Heiga Zen, Ron~J Weiss, Mohammad Norouzi, and William Chan.
\newblock Wavegrad: Estimating gradients for waveform generation.
\newblock \emph{arXiv preprint arXiv:2009.00713}, 2020.

\bibitem[Choi et~al.(2020)Choi, Uh, Yoo, and Ha]{21_choi2020stargan}
Yunjey Choi, Youngjung Uh, Jaejun Yoo, and Jung-Woo Ha.
\newblock Stargan v2: Diverse image synthesis for multiple domains.
\newblock In \emph{Proceedings of the IEEE/CVF conference on computer vision and pattern recognition}, pages 8188--8197, 2020.

\bibitem[Collins et~al.(2020)Collins, Bala, Price, and Susstrunk]{38_collins2020editing}
Edo Collins, Raja Bala, Bob Price, and Sabine Susstrunk.
\newblock Editing in style: Uncovering the local semantics of gans.
\newblock In \emph{Proceedings of the IEEE/CVF Conference on Computer Vision and Pattern Recognition}, pages 5771--5780, 2020.

\bibitem[Deng et~al.(2009)Deng, Dong, Socher, Li, Li, and Fei-Fei]{33_imagenet}
Jia Deng, Wei Dong, Richard Socher, Li-Jia Li, Kai Li, and Li Fei-Fei.
\newblock Imagenet: A large-scale hierarchical image database.
\newblock In \emph{2009 IEEE conference on computer vision and pattern recognition}, pages 248--255. Ieee, 2009.

\bibitem[Denton et~al.(2015)Denton, Chintala, Fergus, et~al.]{10_denton2015deep}
Emily~L Denton, Soumith Chintala, Rob Fergus, et~al.
\newblock Deep generative image models using a laplacian pyramid of adversarial networks.
\newblock \emph{Advances in neural information processing systems}, 28, 2015.

\bibitem[Dhariwal and Nichol(2021)]{30_dhariwal2021diffusion}
Prafulla Dhariwal and Alexander Nichol.
\newblock Diffusion models beat gans on image synthesis.
\newblock \emph{Advances in Neural Information Processing Systems}, 34:\penalty0 8780--8794, 2021.

\bibitem[Goodfellow et~al.(2020)Goodfellow, Pouget-Abadie, Mirza, Xu, Warde-Farley, Ozair, Courville, and Bengio]{09_gan}
Ian Goodfellow, Jean Pouget-Abadie, Mehdi Mirza, Bing Xu, David Warde-Farley, Sherjil Ozair, Aaron Courville, and Yoshua Bengio.
\newblock Generative adversarial networks.
\newblock \emph{Communications of the ACM}, 63\penalty0 (11):\penalty0 139--144, 2020.

\bibitem[H{\"a}rk{\"o}nen et~al.(2020)H{\"a}rk{\"o}nen, Hertzmann, Lehtinen, and Paris]{40_harkonen2020ganspace}
Erik H{\"a}rk{\"o}nen, Aaron Hertzmann, Jaakko Lehtinen, and Sylvain Paris.
\newblock Ganspace: Discovering interpretable gan controls.
\newblock \emph{Advances in Neural Information Processing Systems}, 33:\penalty0 9841--9850, 2020.

\bibitem[Hertz et~al.(2022)Hertz, Mokady, Tenenbaum, Aberman, Pritch, and Cohen-Or]{48_prompt2prompt}
Amir Hertz, Ron Mokady, Jay Tenenbaum, Kfir Aberman, Yael Pritch, and Daniel Cohen-Or.
\newblock Prompt-to-prompt image editing with cross attention control.
\newblock \emph{arXiv preprint arXiv:2208.01626}, 2022.

\bibitem[Ho et~al.(2020)Ho, Jain, and Abbeel]{24_ho2020denoising}
Jonathan Ho, Ajay Jain, and Pieter Abbeel.
\newblock Denoising diffusion probabilistic models.
\newblock \emph{Advances in Neural Information Processing Systems}, 33:\penalty0 6840--6851, 2020.

\bibitem[Ho et~al.(2022)Ho, Saharia, Chan, Fleet, Norouzi, and Salimans]{31_ho2022cascaded}
Jonathan Ho, Chitwan Saharia, William Chan, David~J Fleet, Mohammad Norouzi, and Tim Salimans.
\newblock Cascaded diffusion models for high fidelity image generation.
\newblock \emph{J. Mach. Learn. Res.}, 23\penalty0 (47):\penalty0 1--33, 2022.

\bibitem[Huh et~al.(2020)Huh, Zhang, Zhu, Paris, and Hertzmann]{07_huh2020transforming}
Minyoung Huh, Richard Zhang, Jun-Yan Zhu, Sylvain Paris, and Aaron Hertzmann.
\newblock Transforming and projecting images into class-conditional generative networks.
\newblock In \emph{Computer Vision--ECCV 2020: 16th European Conference, Glasgow, UK, August 23--28, 2020, Proceedings, Part II 16}, pages 17--34. Springer, 2020.

\bibitem[Isola et~al.(2017)Isola, Zhu, Zhou, and Efros]{04_p2p}
Phillip Isola, Jun-Yan Zhu, Tinghui Zhou, and Alexei~A Efros.
\newblock Image-to-image translation with conditional adversarial networks.
\newblock In \emph{Proceedings of the IEEE conference on computer vision and pattern recognition}, pages 1125--1134, 2017.

\bibitem[Jiang et~al.(2021)Jiang, Dai, Wu, and Loy]{62_freq}
Liming Jiang, Bo Dai, Wayne Wu, and Chen~Change Loy.
\newblock Focal frequency loss for image reconstruction and synthesis.
\newblock In \emph{Proceedings of the IEEE/CVF International Conference on Computer Vision}, pages 13919--13929, 2021.

\bibitem[Karras et~al.(2019)Karras, Laine, and Aila]{37_karras2019style}
Tero Karras, Samuli Laine, and Timo Aila.
\newblock A style-based generator architecture for generative adversarial networks.
\newblock In \emph{Proceedings of the IEEE/CVF conference on computer vision and pattern recognition}, pages 4401--4410, 2019.

\bibitem[Kingma et~al.(2021)Kingma, Salimans, Poole, and Ho]{27_kingma2021variational}
Diederik Kingma, Tim Salimans, Ben Poole, and Jonathan Ho.
\newblock Variational diffusion models.
\newblock \emph{Advances in neural information processing systems}, 34:\penalty0 21696--21707, 2021.

\bibitem[Kong et~al.(2020)Kong, Ping, Huang, Zhao, and Catanzaro]{28_kong2020diffwave}
Zhifeng Kong, Wei Ping, Jiaji Huang, Kexin Zhao, and Bryan Catanzaro.
\newblock Diffwave: A versatile diffusion model for audio synthesis.
\newblock \emph{arXiv preprint arXiv:2009.09761}, 2020.

\bibitem[Kumar et~al.(2023)Kumar, Dehghani, and Houlsby]{58_dual}
Manoj Kumar, Mostafa Dehghani, and Neil Houlsby.
\newblock Dual patchnorm.
\newblock \emph{arXiv preprint arXiv:2302.01327}, 2023.

\bibitem[Lee et~al.(2020)Lee, Liu, Wu, and Luo]{22_lee2020maskgan}
Cheng-Han Lee, Ziwei Liu, Lingyun Wu, and Ping Luo.
\newblock Maskgan: Towards diverse and interactive facial image manipulation.
\newblock In \emph{Proceedings of the IEEE/CVF Conference on Computer Vision and Pattern Recognition}, pages 5549--5558, 2020.

\bibitem[Meng et~al.(2021)Meng, Song, Song, Wu, Zhu, and Ermon]{44_sdedit}
Chenlin Meng, Yang Song, Jiaming Song, Jiajun Wu, Jun-Yan Zhu, and Stefano Ermon.
\newblock Sdedit: Image synthesis and editing with stochastic differential equations.
\newblock \emph{arXiv preprint arXiv:2108.01073}, 2021.

\bibitem[Mittal et~al.(2021)Mittal, Engel, Hawthorne, and Simon]{29_mittal2021symbolic}
Gautam Mittal, Jesse Engel, Curtis Hawthorne, and Ian Simon.
\newblock Symbolic music generation with diffusion models.
\newblock \emph{arXiv preprint arXiv:2103.16091}, 2021.

\bibitem[Park et~al.(2019{\natexlab{a}})Park, Liu, Wang, and Zhu]{05_p2phd}
Taesung Park, Ming-Yu Liu, Ting-Chun Wang, and Jun-Yan Zhu.
\newblock Semantic image synthesis with spatially-adaptive normalization.
\newblock In \emph{Proceedings of the IEEE/CVF conference on computer vision and pattern recognition}, pages 2337--2346, 2019{\natexlab{a}}.

\bibitem[Park et~al.(2019{\natexlab{b}})Park, Liu, Wang, and Zhu]{36_spade}
Taesung Park, Ming-Yu Liu, Ting-Chun Wang, and Jun-Yan Zhu.
\newblock Semantic image synthesis with spatially-adaptive normalization.
\newblock In \emph{Proceedings of the IEEE/CVF conference on computer vision and pattern recognition}, pages 2337--2346, 2019{\natexlab{b}}.

\bibitem[Parmar et~al.(2023)Parmar, Singh, Zhang, Li, Lu, and Zhu]{45_p2pzero}
Gaurav Parmar, Krishna~Kumar Singh, Richard Zhang, Yijun Li, Jingwan Lu, and Jun-Yan Zhu.
\newblock Zero-shot image-to-image translation.
\newblock \emph{arXiv preprint arXiv:2302.03027}, 2023.

\bibitem[Patashnik et~al.(2021)Patashnik, Wu, Shechtman, Cohen-Or, and Lischinski]{43_styleclip}
Or Patashnik, Zongze Wu, Eli Shechtman, Daniel Cohen-Or, and Dani Lischinski.
\newblock Styleclip: Text-driven manipulation of stylegan imagery.
\newblock In \emph{Proceedings of the IEEE/CVF International Conference on Computer Vision}, pages 2085--2094, 2021.

\bibitem[Peebles and Xie(2022)]{56_dits}
William Peebles and Saining Xie.
\newblock Scalable diffusion models with transformers.
\newblock \emph{arXiv preprint arXiv:2212.09748}, 2022.

\bibitem[Preechakul et~al.(2022)Preechakul, Chatthee, Wizadwongsa, and Suwajanakorn]{46_diffae}
Konpat Preechakul, Nattanat Chatthee, Suttisak Wizadwongsa, and Supasorn Suwajanakorn.
\newblock Diffusion autoencoders: Toward a meaningful and decodable representation.
\newblock In \emph{Proceedings of the IEEE/CVF Conference on Computer Vision and Pattern Recognition}, pages 10619--10629, 2022.

\bibitem[Pu et~al.(2016)Pu, Gan, Henao, Yuan, Li, Stevens, and Carin]{55_vae}
Yunchen Pu, Zhe Gan, Ricardo Henao, Xin Yuan, Chunyuan Li, Andrew Stevens, and Lawrence Carin.
\newblock Variational autoencoder for deep learning of images, labels and captions.
\newblock \emph{Advances in neural information processing systems}, 29, 2016.

\bibitem[Radford et~al.(2015)Radford, Metz, and Chintala]{11_radford2015unsupervised}
Alec Radford, Luke Metz, and Soumith Chintala.
\newblock Unsupervised representation learning with deep convolutional generative adversarial networks.
\newblock \emph{arXiv preprint arXiv:1511.06434}, 2015.

\bibitem[Radford et~al.(2021)Radford, Kim, Hallacy, Ramesh, Goh, Agarwal, Sastry, Askell, Mishkin, Clark, et~al.]{08_clip}
Alec Radford, Jong~Wook Kim, Chris Hallacy, Aditya Ramesh, Gabriel Goh, Sandhini Agarwal, Girish Sastry, Amanda Askell, Pamela Mishkin, Jack Clark, et~al.
\newblock Learning transferable visual models from natural language supervision.
\newblock In \emph{International conference on machine learning}, pages 8748--8763. PMLR, 2021.

\bibitem[Ramesh et~al.(2022)Ramesh, Dhariwal, Nichol, Chu, and Chen]{01_imagen}
Aditya Ramesh, Prafulla Dhariwal, Alex Nichol, Casey Chu, and Mark Chen.
\newblock Hierarchical text-conditional image generation with clip latents.
\newblock \emph{arXiv preprint arXiv:2204.06125}, 2022.

\bibitem[Rombach et~al.(2022)Rombach, Blattmann, Lorenz, Esser, and Ommer]{02_sd}
Robin Rombach, Andreas Blattmann, Dominik Lorenz, Patrick Esser, and Bj{\"o}rn Ommer.
\newblock High-resolution image synthesis with latent diffusion models.
\newblock In \emph{Proceedings of the IEEE/CVF Conference on Computer Vision and Pattern Recognition}, pages 10684--10695, 2022.

\bibitem[Saharia et~al.(2022{\natexlab{a}})Saharia, Chan, Saxena, Li, Whang, Denton, Ghasemipour, Ayan, Mahdavi, Lopes, et~al.]{03_dalle}
Chitwan Saharia, William Chan, Saurabh Saxena, Lala Li, Jay Whang, Emily Denton, Seyed Kamyar~Seyed Ghasemipour, Burcu~Karagol Ayan, S~Sara Mahdavi, Rapha~Gontijo Lopes, et~al.
\newblock Photorealistic text-to-image diffusion models with deep language understanding.
\newblock \emph{arXiv preprint arXiv:2205.11487}, 2022{\natexlab{a}}.

\bibitem[Saharia et~al.(2022{\natexlab{b}})Saharia, Ho, Chan, Salimans, Fleet, and Norouzi]{32_saharia2022image}
Chitwan Saharia, Jonathan Ho, William Chan, Tim Salimans, David~J Fleet, and Mohammad Norouzi.
\newblock Image super-resolution via iterative refinement.
\newblock \emph{IEEE Transactions on Pattern Analysis and Machine Intelligence}, 2022{\natexlab{b}}.

\bibitem[Salimans et~al.(2016{\natexlab{a}})Salimans, Goodfellow, Zaremba, Cheung, Radford, and Chen]{12_salimans2016improved}
Tim Salimans, Ian Goodfellow, Wojciech Zaremba, Vicki Cheung, Alec Radford, and Xi Chen.
\newblock Improved techniques for training gans.
\newblock \emph{Advances in neural information processing systems}, 29, 2016{\natexlab{a}}.

\bibitem[Salimans et~al.(2016{\natexlab{b}})Salimans, Goodfellow, Zaremba, Cheung, Radford, and Chen]{69_fid}
Tim Salimans, Ian Goodfellow, Wojciech Zaremba, Vicki Cheung, Alec Radford, and Xi Chen.
\newblock Improved techniques for training gans.
\newblock \emph{Advances in neural information processing systems}, 29, 2016{\natexlab{b}}.

\bibitem[Sangkloy et~al.(2017)Sangkloy, Lu, Fang, Yu, and Hays]{35_sangkloy2017scribbler}
Patsorn Sangkloy, Jingwan Lu, Chen Fang, Fisher Yu, and James Hays.
\newblock Scribbler: Controlling deep image synthesis with sketch and color.
\newblock In \emph{Proceedings of the IEEE conference on computer vision and pattern recognition}, pages 5400--5409, 2017.

\bibitem[Schuhmann et~al.(2021)Schuhmann, Vencu, Beaumont, Kaczmarczyk, Mullis, Katta, Coombes, Jitsev, and Komatsuzaki]{34_schuhmann2021laion}
Christoph Schuhmann, Richard Vencu, Romain Beaumont, Robert Kaczmarczyk, Clayton Mullis, Aarush Katta, Theo Coombes, Jenia Jitsev, and Aran Komatsuzaki.
\newblock Laion-400m: Open dataset of clip-filtered 400 million image-text pairs.
\newblock \emph{arXiv preprint arXiv:2111.02114}, 2021.

\bibitem[Shen et~al.(2020)Shen, Gu, Tang, and Zhou]{39_shen2020interpreting}
Yujun Shen, Jinjin Gu, Xiaoou Tang, and Bolei Zhou.
\newblock Interpreting the latent space of gans for semantic face editing.
\newblock In \emph{Proceedings of the IEEE/CVF conference on computer vision and pattern recognition}, pages 9243--9252, 2020.

\bibitem[Shocher et~al.(2020)Shocher, Gandelsman, Mosseri, Yarom, Irani, Freeman, and Dekel]{20_shocher2020semantic}
Assaf Shocher, Yossi Gandelsman, Inbar Mosseri, Michal Yarom, Michal Irani, William~T Freeman, and Tali Dekel.
\newblock Semantic pyramid for image generation.
\newblock In \emph{Proceedings of the IEEE/CVF Conference on Computer Vision and Pattern Recognition}, pages 7457--7466, 2020.

\bibitem[Siarohin et~al.(2019)Siarohin, Lathuili{\`e}re, Tulyakov, Ricci, and Sebe]{54_fomm}
Aliaksandr Siarohin, St{\'e}phane Lathuili{\`e}re, Sergey Tulyakov, Elisa Ricci, and Nicu Sebe.
\newblock First order motion model for image animation.
\newblock \emph{Advances in Neural Information Processing Systems}, 32, 2019.

\bibitem[Sohl-Dickstein et~al.(2015)Sohl-Dickstein, Weiss, Maheswaranathan, and Ganguli]{23_sohl2015deep}
Jascha Sohl-Dickstein, Eric Weiss, Niru Maheswaranathan, and Surya Ganguli.
\newblock Deep unsupervised learning using nonequilibrium thermodynamics.
\newblock In \emph{International Conference on Machine Learning}, pages 2256--2265. PMLR, 2015.

\bibitem[Song et~al.(2020)Song, Sohl-Dickstein, Kingma, Kumar, Ermon, and Poole]{25_song2020score}
Yang Song, Jascha Sohl-Dickstein, Diederik~P Kingma, Abhishek Kumar, Stefano Ermon, and Ben Poole.
\newblock Score-based generative modeling through stochastic differential equations.
\newblock \emph{arXiv preprint arXiv:2011.13456}, 2020.

\bibitem[Tewari et~al.(2020)Tewari, Elgharib, Bharaj, Bernard, Seidel, P{\'e}rez, Zollhofer, and Theobalt]{41_tewari2020stylerig}
Ayush Tewari, Mohamed Elgharib, Gaurav Bharaj, Florian Bernard, Hans-Peter Seidel, Patrick P{\'e}rez, Michael Zollhofer, and Christian Theobalt.
\newblock Stylerig: Rigging stylegan for 3d control over portrait images.
\newblock In \emph{Proceedings of the IEEE/CVF Conference on Computer Vision and Pattern Recognition}, pages 6142--6151, 2020.

\bibitem[Vaswani et~al.(2017)Vaswani, Shazeer, Parmar, Uszkoreit, Jones, Gomez, Kaiser, and Polosukhin]{57_vits}
Ashish Vaswani, Noam Shazeer, Niki Parmar, Jakob Uszkoreit, Llion Jones, Aidan~N Gomez, {\L}ukasz Kaiser, and Illia Polosukhin.
\newblock Attention is all you need.
\newblock \emph{Advances in neural information processing systems}, 30, 2017.

\bibitem[Wu et~al.(2021)Wu, Lischinski, and Shechtman]{42_wu2021stylespace}
Zongze Wu, Dani Lischinski, and Eli Shechtman.
\newblock Stylespace analysis: Disentangled controls for stylegan image generation.
\newblock In \emph{Proceedings of the IEEE/CVF Conference on Computer Vision and Pattern Recognition}, pages 12863--12872, 2021.

\bibitem[Yi et~al.(2020)Yi, Tang, Azizi, Jang, and Xu]{16_yi2020contextual}
Zili Yi, Qiang Tang, Shekoofeh Azizi, Daesik Jang, and Zhan Xu.
\newblock Contextual residual aggregation for ultra high-resolution image inpainting.
\newblock In \emph{Proceedings of the IEEE/CVF conference on computer vision and pattern recognition}, pages 7508--7517, 2020.

\bibitem[Zhang et~al.(2020)Zhang, Zhang, Chen, Yuan, and Wen]{18_zhang2020cross}
Pan Zhang, Bo Zhang, Dong Chen, Lu Yuan, and Fang Wen.
\newblock Cross-domain correspondence learning for exemplar-based image translation.
\newblock In \emph{Proceedings of the IEEE/CVF Conference on Computer Vision and Pattern Recognition}, pages 5143--5153, 2020.

\bibitem[Zhao et~al.(2016)Zhao, Mathieu, and LeCun]{13_zhao2016energy}
Junbo Zhao, Michael Mathieu, and Yann LeCun.
\newblock Energy-based generative adversarial network.
\newblock \emph{arXiv preprint arXiv:1609.03126}, 2016.

\bibitem[Zhao et~al.(2020)Zhao, Mo, Lin, Wang, Zuo, Chen, Xing, and Lu]{15_zhao2020uctgan}
Lei Zhao, Qihang Mo, Sihuan Lin, Zhizhong Wang, Zhiwen Zuo, Haibo Chen, Wei Xing, and Dongming Lu.
\newblock Uctgan: Diverse image inpainting based on unsupervised cross-space translation.
\newblock In \emph{Proceedings of the IEEE/CVF conference on computer vision and pattern recognition}, pages 5741--5750, 2020.

\bibitem[Zheng et~al.(2019)Zheng, Cham, and Cai]{14_zheng2019pluralistic}
Chuanxia Zheng, Tat-Jen Cham, and Jianfei Cai.
\newblock Pluralistic image completion.
\newblock In \emph{Proceedings of the IEEE/CVF Conference on Computer Vision and Pattern Recognition}, pages 1438--1447, 2019.

\bibitem[Zhu et~al.(2016)Zhu, Kr{\"a}henb{\"u}hl, Shechtman, and Efros]{06_zhu2016generative}
Jun-Yan Zhu, Philipp Kr{\"a}henb{\"u}hl, Eli Shechtman, and Alexei~A Efros.
\newblock Generative visual manipulation on the natural image manifold.
\newblock In \emph{Computer Vision--ECCV 2016: 14th European Conference, Amsterdam, The Netherlands, October 11-14, 2016, Proceedings, Part V 14}, pages 597--613. Springer, 2016.

\end{thebibliography}
}

\ifarxiv \clearpage \appendix \section{Appendix}
\label{sec:appendix} \fi

\end{document}